\documentclass{article}

\usepackage{PRIMEarxiv}
\usepackage{tabularx}
\usepackage{booktabs}
\usepackage{hyperref}
\usepackage{listings}
\usepackage{xcolor}
\usepackage{dsfont}
\usepackage{graphicx}
\usepackage{amsmath, amsfonts}
\usepackage{cite}
\usepackage{array}
\usepackage{microtype}
\usepackage{subcaption}
\usepackage[capitalize]{cleveref}
\lstdefinelanguage{ASP}{
  morecomment=[l]{\%},
  morestring=[b]",
  sensitive=true,
  morekeywords={
    not
  }
}

\lstdefinelanguage{ASP}{
  morecomment=[l]{\%},
  morestring=[b]",
  sensitive=true,
  morekeywords={
    not
  }
}
\title{PACE: A Neuro-Symbolic Framework for Plausible and Actionable Counterfactual Explanations}

\author{
Pavel Iakovets$^{1}$\thanks{Corresponding author: \texttt{paiakovets@edu.aau.at}} \and
Liyanapathiranage Sudeepika Wajirakumari Samarathunga$^{2}$ \and
Martin Thomas Horsch$^{2}$ \and
Fadi Al Machot$^{2}$\\[1ex]
$^{1}$University of Klagenfurt, Universitätsstraße 65/67, 9020 Klagenfurt am Wörthersee, Austria\\
$^{2}$Norwegian University of Life Sciences, Elizabeth Stephansens v. 15, 1433 Ås, Norway
}

\date{}

\begin{document}

\maketitle
\begin{abstract}
Counterfactual explanations have emerged as a practical approach for explaining machine learning predictions by identifying minimal changes to an input instance that would alter a model's decision. Although many existing methods successfully generate prediction-changing alternatives, they often produce unrealistic or infeasible recommendations because they lack explicit mechanisms for incorporating domain knowledge and intervention constraints. Neuro-symbolic artificial intelligence offers a promising direction by combining data-driven predictive models with symbolic reasoning capable of representing human-understandable rules and feasible actions.
This paper presents PACE, a modular neuro-symbolic framework for generating feasibility-aware counterfactual explanations. The framework separates prediction and reasoning into two complementary components: a neural predictive model responsible for classification and a symbolic reasoning layer that enforces domain-specific constraints during counterfactual generation. By explicitly modeling feasible interventions, the framework aims to produce explanations that are more consistent with domain knowledge while remaining interpretable and actionable. The proposed approach is model-agnostic and can be adapted to application domains where realistic decision support is required.
To demonstrate the applicability of the framework, a case study is conducted on the Adult Income dataset. A multilayer perceptron classifier is combined with Answer Set Programming (ASP) rules that encode feasible modifications to education, occupation, and working hours while preserving immutable attributes. Experimental results highlight the trade-off between counterfactual validity and plausibility and show that incorporating symbolic constraints yields explanations that better satisfy domain-specific feasibility requirements. These findings illustrate the potential of neuro-symbolic methods for supporting the generation of transparent, feasibility-aware counterfactual explanations in explainable AI systems.

\end{abstract}

%\begin{keywords}
%Neuro-Symbolic AI \sep
%Counterfactual Explanations \sep
%Explainable Artificial Intelligence (XAI) \sep
%Answer Set Programming \sep
%Algorithmic Recourse \sep
%Knowledge-Based Reasoning
%\end{keywords}

\maketitle
\section{Introduction}

The increasing use of machine learning models in decision-support systems has intensified the demand for explanations that are transparent, actionable, and trustworthy~\cite{rudin2022interpretable,stepin2021survey,karimi2022survey}. Among the many approaches proposed in Explainable Artificial Intelligence (XAI), counterfactual explanations have attracted particular attention because they provide actionable recommendations by identifying how an instance could be modified to obtain a different prediction~\cite{miller2019explanation}. Such explanations are especially valuable in domains such as healthcare, finance, education, and public administration, where decisions directly affect individuals.

Despite their appeal, generating useful counterfactual explanations remains challenging. A desirable counterfactual should be \emph{valid}, \emph{minimal}, and \emph{plausible}. However, many existing methods primarily optimize for prediction change and proximity, often producing recommendations that violate domain constraints or cannot realistically be implemented~\cite{mahajan2019preserving}. As a result, there is growing interest in approaches that incorporate domain knowledge directly into the explanation process, enabling explanations that are not only valid but also feasible in practice.

Neuro-Symbolic Artificial Intelligence (NeSy) provides a promising foundation for addressing this challenge by combining the representation-learning capabilities of neural networks with the reasoning capabilities of symbolic systems~\cite{garcez2023neurosymbolic}. Recent work has demonstrated the value of neuro-symbolic methods for integrating logical constraints, fairness requirements, and domain knowledge into machine learning systems~\cite{heilmann2026neurosymbolic}. These characteristics make neuro-symbolic reasoning a natural candidate for feasibility-aware counterfactual explanation generation.

This work investigates how symbolic knowledge can be incorporated into the generation of counterfactual explanations through explicit representations of admissible interventions and domain constraints. We formulate counterfactual generation as a constrained search problem in which candidate explanations must satisfy both predictive objectives and symbolic feasibility requirements. To explore this idea, we develop PACE, a neuro-symbolic framework that combines machine learning prediction with symbolic reasoning over a domain-specific knowledge base.

A case study is conducted on the Adult Income dataset using a multilayer perceptron (MLP) classifier and Answer Set Programming (ASP) for symbolic reasoning. The proposed framework is compared with representative counterfactual generation approaches from different paradigms, including Random Search, Diverse Counterfactual Explanations (DiCE)~\cite{mothilal2020explaining}, Wachter~\cite{wachter2017counterfactual}, Variational Counterfactual Network (VCNet)~\cite{guyomard2022vcnet}, and Counterfactual Conditional Heterogeneous Variational Autoencoder (C-CHVAE)~\cite{pawelczyk2020learning}. The results highlight the trade-off between validity and plausibility in counterfactual generation and demonstrate the effect of explicitly enforcing symbolic feasibility constraints during the search process.

The main contributions of this work are as follows:

\begin{enumerate}
\item A modular neuro-symbolic framework that integrates machine learning models with symbolic reasoning for constrained counterfactual explanation generation.

\item A feasibility-aware formulation of counterfactual generation in which symbolic constraints explicitly define the admissible intervention space.

\item An ASP-based implementation that enforces domain-specific feasibility constraints during counterfactual generation.

\item An empirical evaluation on the Adult Income dataset comparing the proposed framework with Random Search, DiCE, Wachter, VCNet, and C-CHVAE, highlighting the trade-off between validity and plausibility.

\end{enumerate}

The remainder of this paper is organized as follows. Section~\ref{sec:related_work} reviews related work. Sections~3--5 present the proposed framework, architecture, and implementation. Section~6 reports the experimental results, Section~7 discusses the findings and limitations, and Section~8 concludes the paper.

\section{Related Work}
\label{sec:related_work}

Existing counterfactual generation methods can be broadly grouped into three categories. The first category comprises optimization-based approaches, such as the method of Wachter et al.~\cite{wachter2017counterfactual} and DiCE~\cite{mothilal2020explaining}. These methods search for prediction-changing instances while minimizing changes to the original input and, in the case of DiCE, encouraging diversity among the generated explanations. Although effective at finding counterfactuals, feasibility is typically handled as a soft constraint rather than being explicitly guaranteed.
The second category includes generative approaches that learn the data manifold and generate realistic candidate explanations. Representative examples include C-CHVAE~\cite{pawelczyk2020learning}, which searches in a latent representation space, and more recent generative methods such as VCNet~\cite{guyomard2022vcnet}. These approaches improve realism by constraining counterfactuals to remain close to the data distribution, but they generally rely on statistical regularities rather than explicit domain knowledge.
A third line of research incorporates symbolic reasoning and domain constraints into counterfactual generation. Bertossi and Reyes~\cite{bertossi2021answer} demonstrate that answer-set programs can be used to declaratively specify counterfactual interventions on entities under classification, reason about them, and include domain knowledge through query answering. More broadly, neuro-symbolic AI combines the representation-learning capabilities of neural networks with the transparency and reasoning capabilities of symbolic systems~\cite{acharya2025comprehensive}. Recent work has integrated counterfactual fairness into the neurosymbolic framework of Logic Tensor Networks (LTN), demonstrating advantages including intrinsic interpretability and flexibility in handling subgroup fairness~\cite{heilmann2026neurosymbolic}.
Despite these advances, a gap remains between high-validity counterfactual generation methods and approaches that explicitly enforce feasibility. Optimization-based and generative techniques often treat actionability as a secondary objective, while symbolic approaches frequently focus on intervention analysis rather than explanation generation for predictive models. Consequently, many existing methods may generate counterfactuals that successfully change predictions but violate domain-specific requirements.

\section{Methodology}

\subsection{Problem Formulation}
\label{subsec:problem_formulation}
We formalize the generation of a plausible counterfactual explanation as a constrained search problem over a domain-restricted input space. Let $f: \mathcal{X} \to \mathcal{Y}$ denote a trained, black-box predictive model that maps an input instance $\mathbf{x} \in \mathcal{X}$ to a discrete categorical outcome $y \in \mathcal{Y}$. Given a specific factual instance $\mathbf{x}$ for which the model yields an initial prediction $f(\mathbf{x}) = y$, the objective is to find a counterfactual instance $\mathbf{x}^{cf}$ that alters the model's classification to a target outcome $y^{cf} \neq y$. 

To ensure the explanation is practical for an end user, the transition from $\mathbf{x}$ to $\mathbf{x}^{cf}$ must minimize a distance metric that quantifies the magnitude of the applied modifications while remaining strictly within the set of valid states. Formally, the counterfactual explanation $\mathbf{x}^{cf}$ is defined as the solution to the following constrained formulation:

\begin{equation}
\mathbf{x}^{cf} = \arg\min_{\mathbf{x}' \in \Omega} d(\mathbf{x}, \mathbf{x}')
\label{eq:cf_objective}
\end{equation}

\noindent subject to the prediction-flip condition:

\begin{equation}
f(\mathbf{x}^{cf}) \neq f(\mathbf{x})
\label{eq:prediction_flip}
\end{equation}

\noindent where $\Omega \subseteq \mathcal{X}$ represents the feasible intervention space determined by domain-specific rules. In this work, the distance $d$ is instantiated as the number of modified
features,
\begin{equation}
d(\mathbf{x}, \mathbf{x}') = \sum_{j} \mathds{1}\left[x_j \neq x'_j\right],
\label{eq:distance}
\end{equation}
so that minimizing $d$ corresponds to minimizing the number of interventions
(the minimality criterion). The budget $K$ in the search procedure directly
bounds this count.
In this formulation, the predictive model $f$ is treated as an oracle that evaluates candidates generated within the valid subspace $\Omega$. The objective separates the verification of semantic consistency, which is handled implicitly via restriction to $\Omega$, from the minimization of feature modifications, which is handled by the search metric $d$.

\subsection{Symbolic Constraint Modeling}
\label{subsec:symbolic_constraint}
The feasible intervention space $\Omega$ is defined by a declarative symbolic knowledge base that encodes background information, logic rules, and structural dependencies governing the domain attributes. In practical explanation contexts, features cannot be modified uniformly or independently. We partition the feature space into immutable attributes $\mathcal{X}_I$ and editable attributes $\mathcal{X}_E$, such that $\mathcal{X} = \mathcal{X}_I \cup \mathcal{X}_E$ and $\mathcal{X}_I \cap \mathcal{X}_E = \emptyset$. Immutable attributes represent inherent properties that cannot be changed by user intervention or environmental drift, enforcing a hard constraint where $\mathbf{x}'_i = \mathbf{x}_i$ for all $i \in \mathcal{X}_I$. Editable attributes represent features that are susceptible to modification, subject to directional constraints and relational dependencies.

The symbolic knowledge base maps these attributes to logical predicates and evaluates candidates against a system of constraints. Let $\mathcal{R} = \{R_1, R_2, \dots, R_m\}$ be a set of symbolic rules representing domain knowledge. A rule can represent a directional invariant, preventing a feature from changing in an impossible direction, or a relational invariant, specifying that a change in one feature necessitates a corresponding change in another. The feasible space is therefore defined as:

\begin{equation}
\Omega = \{ \mathbf{x}' \in \mathcal{X} \mid \forall i \in \mathcal{X}_I, \mathbf{x}'_i = \mathbf{x}_i \;\wedge\; \forall R_k \in \mathcal{R}, \mathbf{x}' \models R_k \}
\label{eq:feasible_space}
\end{equation}

\noindent where $\mathbf{x}' \models R_k$ denotes that the candidate instance satisfies the logical constraint $R_k$. By utilizing a symbolic reasoning engine to construct and evaluate Equation~\ref{eq:feasible_space}, the framework eliminates invalid areas of the unconstrained space $\mathcal{X}$. This symbolic filtering restricts the downstream evaluation exclusively to candidates that respect the causal and logical axioms of the application domain.

\subsection{Counterfactual Search Procedure}
\label{subsec:search_procedure}
The framework operates via an iterative generate-and-verify search loop that systematically expands the intervention distance until an optimal prediction-changing instance is identified. The procedure begins by translating the factual instance $\mathbf{x}$ into a set of symbolic facts. The symbolic reasoner utilizes these facts along with the rule base $\mathcal{R}$ to establish the immutable baseline and map out valid paths of modification within the editable subspace $\mathcal{X}_E$.

The search space is structured incrementally based on an explicit budget constraint $K$, which restricts the maximum number of simultaneous feature alterations or the maximum allowable distance metric value. At each iteration, the symbolic engine acts as a generator, producing a set of candidate interventions that satisfy all constraints in $\Omega$ within the current budget $K$. Because the framework prioritizes sparse explanations, the search starts with a minimal budget, targeting single-attribute modifications before exploring multi-attribute combinations.

Once a batch of feasible candidates is generated, they are mapped back into the format required by the predictive model. The model $f$ performs a forward evaluation on each candidate to check the prediction-flip condition specified in Equation~\ref{eq:prediction_flip}. If a single candidate or multiple candidates satisfy the condition, the search terminates, and the framework returns the candidate $\mathbf{x}^{cf}$ that minimizes $d(\mathbf{x}, \mathbf{x}')$. If none of the generated candidates succeed in flipping the prediction of $f$, the search budget $K$ is incrementally increased, expanding the search radius within $\Omega$ to allow for broader or more complex attribute modifications. This process repeats until a valid counterfactual is found or the maximum search bound is reached, ensuring that the returned explanation represents the sparsest possible intervention that achieves the target classification.

\section{Case Study: Counterfactual Explanations for Income Classification}
\label{sec:case_study}

To demonstrate the practical applicability of the proposed neuro-symbolic framework, we consider a counterfactual explanation task based on the Adult Income dataset \cite{adult_2}. The objective is to generate feasible and interpretable counterfactual explanations for a binary income classification model while respecting domain-specific constraints on feature modifications.

The Adult Income dataset is a widely used benchmark for tabular classification and explainable artificial intelligence (XAI) research. The prediction task consists of determining whether an individual's annual income exceeds \$50,000 based on demographic and employment-related attributes. Following data cleaning and the removal of records containing missing categorical values, the resulting dataset contains 30,718 instances. The target variable is represented as a binary class indicating whether income is less than or equal to \$50K or greater than \$50K. 

The dataset is partitioned using a stratified train-test split, yielding 24,574 training instances and 6,144 testing instances. Stratification preserves the original class distribution, where approximately 75\% of instances belong to the lower-income category. To maintain interpretability and facilitate symbolic reasoning, only a subset of five features is used in this case study, as detailed in Table~\ref{tab:case_features}. These features represent different attribute types commonly encountered in counterfactual generation problems, including immutable personal characteristics, categorical variables, and continuous variables.

\begin{table}[ht]
\centering
\caption{Features used in the case study.}
\label{tab:case_features}
\begin{tabular}{lll}
\toprule
Feature & Type & Role \\
\midrule
Age & Numerical & Immutable \\
Sex & Categorical & Immutable \\
Education & Categorical & Editable \\
Occupation & Categorical & Editable \\
Hours per Week & Numerical & Editable \\
\bottomrule
\end{tabular}
\end{table}

Numerical attributes are standardized prior to model training, while categorical variables are transformed using one-hot encoding. The symbolic component, however, operates on the original human-readable feature values to preserve interpretability during counterfactual generation.

\subsection{Predictive Model}

The neural component of the framework is implemented as a multilayer perceptron (MLP) classifier. The model is trained once and subsequently treated as a fixed predictive function during counterfactual generation; model parameters remain unchanged throughout the explanation process.

Formally, let 
\begin{equation}
f_{\theta} : \mathcal{X} \rightarrow \mathcal{Y}
\end{equation}
denote the trained classifier parameterized by $\theta$, where $\mathcal{X}$ is the feature space and $\mathcal{Y}=\{0,1\}$ represents the two income classes. Given an instance $\mathbf{x} \in \mathcal{X}$, the model prediction is defined as
\begin{equation}
y = f_{\theta}(\mathbf{x}).
\end{equation}
The structural and training configuration of this predictive model is summarized in Table~\ref{tab:mlp_config}. The classification accuracy is 0.82.

\begin{table}[ht]
\centering
\caption{Configuration of the predictive model.}
\label{tab:mlp_config}
\begin{tabular}{ll}
\toprule
Component & Configuration \\
\midrule
Model & Multilayer Perceptron \\
Input Features & 5 \\
Hidden Layers & 2 \\
Layer Sizes & 32, 16 \\
Activation & ReLU \\
Optimizer & Adam \\
Learning Rate & 0.001 \\
Batch Size & 128 \\
Regularization & $5\times10^{-4}$ \\
Maximum Epochs & 120 \\
Early Stopping & Enabled \\
\bottomrule
\end{tabular}
\end{table}

\subsection{Symbolic Knowledge Base}

The symbolic component defines the space of admissible interventions. Rather than allowing arbitrary feature modifications, the framework incorporates domain knowledge through explicit symbolic constraints. These constraints determine which attributes may be modified and how such modifications may occur.

Let $\mathcal{C} = \{c_1,c_2,\ldots,c_m\}$ denote the set of symbolic constraints. The feasible intervention space $\Omega$ is defined as
\begin{equation}
\Omega = \left\{ \mathbf{x}' \;\middle|\; c_j(\mathbf{x}') = 1, \forall c_j \in \mathcal{C} \right\}.
\end{equation}
Here, $\mathcal{C}$ is the case-study instantiation of the general rule
base $\mathcal{R}$ defined in Section~3, specialized to the constraints of the Adult Income task.
As a result, only candidate counterfactuals belonging to $\Omega$ are considered valid during the search process. The concrete symbolic rules used in this case study are summarized in Table~\ref{tab:symbolic_constraints}.

\begin{table}[ht]
\centering
\caption{Symbolic constraints used for counterfactual generation.}
\label{tab:symbolic_constraints}
\begin{tabular}{ll}
\toprule
Constraint & Description \\
\midrule
Age & Fixed attribute \\
Sex & Fixed attribute \\
Education & Adjacent-level transitions only \\
Occupation & Predefined transition graph \\
Hours per Week & Bounded step modifications \\
\bottomrule
\end{tabular}
\end{table}

The education attribute is modeled as an ordered progression structure in which transitions are allowed only between neighboring educational levels. Similarly, occupation changes are restricted to predefined transition pairs representing plausible career movements. Weekly working hours may increase or decrease only within a limited range, preventing unrealistic recommendations (see Appendix for additional details).

\subsection{Counterfactual Search Process}
For the income classification task, counterfactual generation follows the constrained formulation introduced in Section~3 (Equations~\ref{eq:cf_objective} and \ref{eq:prediction_flip}): given an original instance $\mathbf{x}$ and its prediction $f_{\theta}(\mathbf{x})$, we search for the closest instance $\mathbf{x}^{cf} \in \Omega$ that flips the prediction, where $\Omega$ is the feasible intervention space defined by the symbolic knowledge base of Section~ 4.2.

The search procedure follows a budget-based strategy. Candidate explanations are first generated with the smallest possible intervention budget. If no prediction-changing counterfactual is found, the budget is gradually increased. This process prioritizes sparse explanations that require only a small number of feature modifications.

\subsection{Experimental Setup}

The proposed framework is compared against several representative counterfactual generation approaches. DiCE is evaluated using its official implementation, while Wachter-style, VCNet-style, and C-CHVAE-style baselines are simplified implementations designed to capture the core search principles of their respective methods rather than exact reproductions of the original algorithms. An unconstrained random-search baseline is also included for comparison. All methods operate under comparable evaluation settings on the same test instances.
The evaluation focuses on five complementary performance metrics:

\begin{description}
    \item[\textbf{Validity}:] The proportion of instances for which a prediction-changing counterfactual is successfully discovered.
    \item[\textbf{Minimality}:] The average number of features modified across the generated counterfactuals.
    \item[\textbf{Plausibility}:] The proportion of generated explanations that strictly satisfy the defined symbolic constraints.
    \item[\textbf{Runtime}:] The average computational time required to generate an explanation per instance.
    \item[\textbf{Iterations to Flip}:] The average number of candidate evaluations required before a valid prediction change is achieved.
\end{description}

This case study provides a concrete instantiation of the proposed neuro-symbolic framework, enabling an empirical investigation into the trade-offs between counterfactual validity and symbolic plausibility in realistic decision-support scenarios.
\section{Results}

The experimental evaluation compares the proposed PACE framework against five baselines: an unconstrained Random Search strategy, DiCE, a Wachter-style optimization method, and two generative approaches (VCNet-style and C-CHVAE-style). A total of 1,000 unseen test instances were evaluated. The quantitative results are summarized in Table~\ref{tab:cf_comparison}.

Table~\ref{tab:cf_comparison} reveals a clear trade-off between validity and plausibility in counterfactual generation. Random Search and DiCE achieve the highest validity scores (0.778 and 0.742, respectively), indicating a greater ability to identify prediction-changing interventions. However, both methods exhibit very low plausibility scores (0.027 and 0.042), suggesting that many generated counterfactuals violate the domain-specific feasibility constraints defined for this case study.

In contrast, PACE achieves perfect plausibility by restricting the search process to interventions that satisfy the symbolic constraints. As a result, all generated explanations remain consistent with the admissible intervention space defined by the knowledge base. PACE also generates relatively compact counterfactuals, requiring 1.242 feature modifications on average, fewer than Random Search, DiCE, VCNet-style, and C-CHVAE-style methods.

Among the baseline methods, VCNet-style achieves the strongest combination of validity and plausibility, while C-CHVAE-style provides intermediate performance. The Wachter-style baseline achieves the lowest validity and plausibility scores, reflecting the difficulty of unconstrained optimization in producing feasible interventions within this domain.

As a descriptive summary of the trade-off between validity and plausibility, Table~\ref{tab:cf_comparison} also reports the product of these two metrics. This composite metric is reported only as a descriptive summary of the observed trade-off. Under this criterion, PACE achieves the highest score (0.240), reflecting its strong compliance with the symbolic constraints despite lower validity than unconstrained approaches.

\begin{table*}[t]
\centering
\caption{Comparison of counterfactual generation methods on 1,000 Adult Income test instances.}
\label{tab:cf_comparison}
\begin{tabularx}{\textwidth}{l l *{6}{>{\centering\arraybackslash}X}}
\toprule
\textbf{Method} & \textbf{Implementation} & \textbf{Validity} & \textbf{Minimality}  & \textbf{Plausibility} & \textbf{Validity $\times$ Plausibility}  & \textbf{Avg. Iterations to flip} & \textbf{Avg. Runtime (s)} \\
\midrule
PACE (Ours)     & Own     & 0.240          & 1.242& \textbf{1.000} & \textbf{0.240} & 5.271          & 0.076 \\
Wachter-style   & Own     & 0.131          & \textbf{1.023 }         & 0.008          & 0.001          & 83.557         & 2.127 \\
VCNet-style     & Own     & 0.325          & 1.365          & 0.452          & 0.146          & 37.105         & 0.060 \\
C-CHVAE-style   & Own     & 0.531          & 1.527          & 0.171          & 0.091          & 1.307          & \textbf{0.029} \\
DiCE            & Library & 0.742          & 1.953          & 0.042          & 0.031          & \textbf{1.000} & 0.094 \\
Random Baseline & Own     & \textbf{0.778} & 1.671          & 0.027          & 0.021          & 31.400         & 0.055 \\
\bottomrule
\end{tabularx}
\end{table*}

\section{Discussion and Additional Analysis}

Plausibility in this study is defined with respect to the symbolic constraints encoded in the knowledge base. Consequently, the perfect plausibility achieved by PACE is expected by construction. To separate the contribution of the constraints from that of symbolic search, we evaluated all methods on the same feasible intervention space $\Omega$.

\begin{table}[ht]
\centering
\caption{Matched-constraint comparison on 200 Adult Income test instances
(budget $K{=}2$, $500$ candidates per instance). All three methods are
restricted to the same feasible set $\Omega$: \emph{blind + reject} proposes
constraint-unaware candidates and discards infeasible ones, \emph{search in
$\Omega$} samples only feasible candidates, and \emph{PACE} enumerates $\Omega$ exhaustively. \emph{Feasible rate} is the fraction of proposals lying in $\Omega$. Plausibility is $1.0$ for all methods by construction once they share $\Omega$.}
\label{tab:matched}
\begin{tabular}{lccccc}
\toprule
Method & Validity & Minimality & Plausibility & Feasible rate & Runtime (s) \\
\midrule
Random (blind + reject)       & 0.220 & 1.091 & 1.000 & 0.108 & 0.067 \\
Random (search in $\Omega$)   & 0.295 & 1.305 & 1.000 & 1.000 & 0.066 \\
PACE (ASP search in $\Omega$) & 0.295 & 1.305 & 1.000 & 1.000 & 0.097 \\
\bottomrule
\end{tabular}
\end{table}

As shown in Table~\ref{tab:matched}, plausibility becomes identical once all methods share the same feasible set, and random search restricted to $\Omega$ achieves the same validity and minimality as PACE. This suggests that the plausibility gains primarily arise from explicit constraint specification, while the low feasible rate of unconstrained search explains the poor compliance of unconstrained baselines.

\begin{figure}[ht]
\centering
\includegraphics[width=0.85\linewidth]{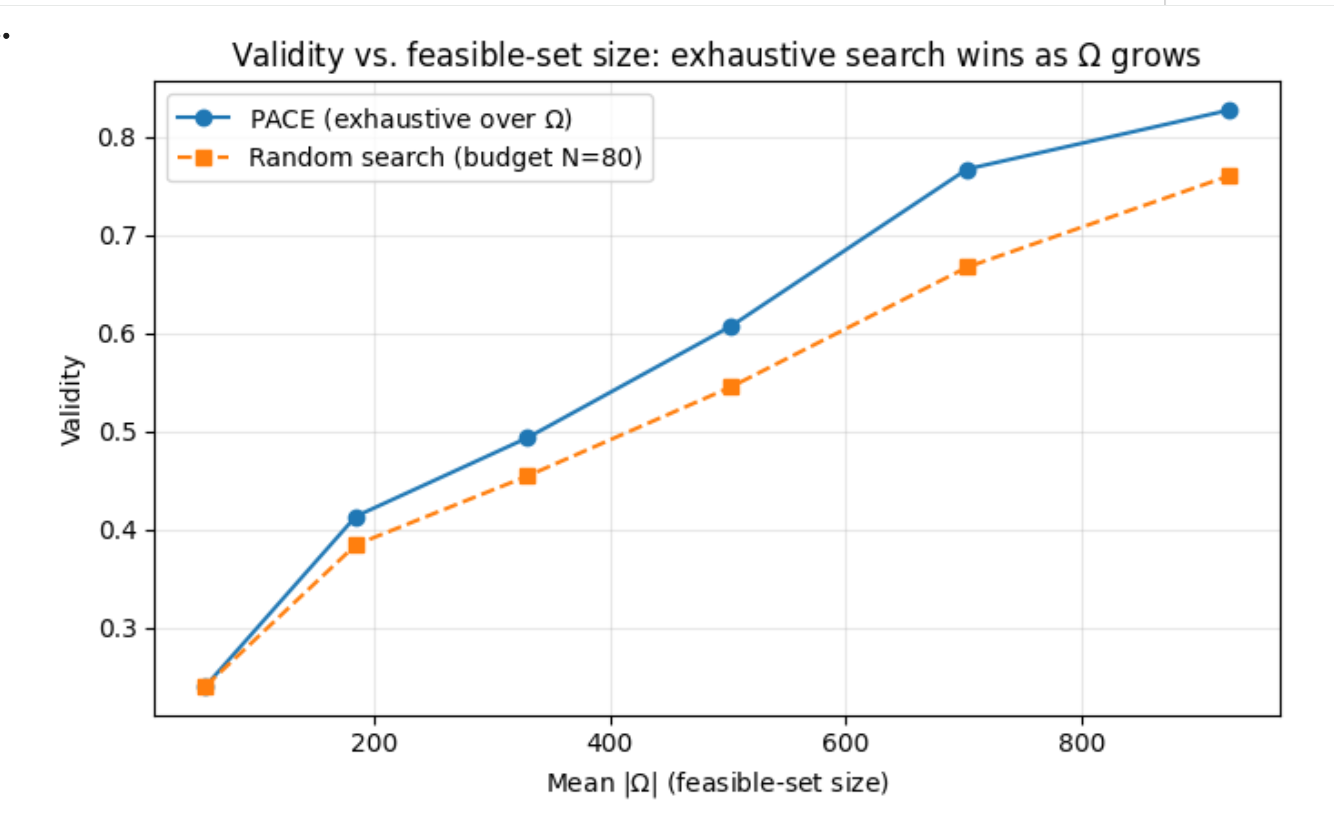}
\caption{Validity versus feasible-set size $|\Omega|$.}
\label{fig:omega}
\end{figure}

The contribution of symbolic search becomes apparent when the feasible intervention space grows. Figure~\ref{fig:omega} and Table~\ref{tab:scaling} compare exhaustive search with random sampling under a fixed candidate budget.

\begin{table}[ht]
\centering
\caption{Validity as the feasible set grows, for exhaustive search (a verified proxy for PACE) versus random search with a fixed budget of $N{=}80$ candidates. The feasible set is enlarged by widening the admissible transition neighbourhoods to graph distance $w$; \emph{Mean $|\Omega|$} is the resulting average feasible-set size and \emph{Gap} is the validity difference (exhaustive $-$ random). The $w{=}1$ row corresponds to the base constraints of Table~\ref{tab:cf_comparison}; it is computed here on the scalability sample and is not directly comparable to the 200-instance matched run in Table~\ref{tab:matched}.}
\label{tab:scaling}
\begin{tabular}{ccccc}
\toprule
Width $w$ & Mean $|\Omega|$ & Exhaustive & Random & Gap \\
\midrule
1 & 57.4  & 0.240 & 0.240 & 0.000 \\
2 & 184.7 & 0.413 & 0.384 & 0.029 \\
3 & 330.0 & 0.493 & 0.454 & 0.039 \\
4 & 502.9 & 0.607 & 0.545 & 0.062 \\
5 & 702.9 & 0.767 & 0.667 & 0.100 \\
6 & 924.6 & 0.827 & 0.760 & 0.066 \\
\bottomrule
\end{tabular}
\end{table}

For small feasible sets, exhaustive and random search perform similarly. As $|\Omega|$ increases, exhaustive search achieves higher validity, indicating that symbolic search contributes completeness through systematic exploration of the admissible intervention space.

Finally, we evaluated plausibility using an external criterion independent of the symbolic rules. The mean distance to the $k=5$ nearest training neighbours was $0.354$ for PACE counterfactuals and $0.944$ for unconstrained search, indicating that the imposed constraints produce counterfactuals substantially closer to the empirical data manifold.

Several limitations should be acknowledged. The evaluation was conducted on a single benchmark dataset and relied on manually specified constraints. Future work should investigate additional datasets, richer forms of domain knowledge, automated constraint acquisition, and more complex relational and temporal constraints.

\section{Conclusion}

This paper presented PACE, a neuro-symbolic framework for generating feasibility-aware counterfactual explanations through the integration of machine learning and symbolic reasoning. Counterfactual generation was formulated as a constrained search problem in which candidate explanations must satisfy both predictive objectives and domain-specific intervention constraints.

The framework was evaluated on the Adult Income dataset using a multilayer perceptron classifier and Answer Set Programming. A comparison with Random Search, DiCE, Wachter-style, VCNet-style, and C-CHVAE-style baselines on 1,000 test instances revealed a clear trade-off between validity and plausibility. While unconstrained and generative approaches achieved higher validity, they frequently produced explanations that violated the symbolic feasibility constraints. In contrast, PACE generated counterfactuals that satisfied all specified constraints and required relatively few feature modifications on average, reflecting the explicit enforcement of feasibility requirements during the search process.

Future work will explore richer forms of symbolic knowledge, automated constraint acquisition, and evaluation across additional datasets and application domains.

\bibliographystyle{unsrt}  
\bibliography{references}

\appendix

\section{ASP Knowledge Base for Counterfactual Generation}
\label{app:asp}

The complete Answer Set Programming (ASP) knowledge base used in the PACE framework is reported in Listing~\ref{lst:asp}. The program encodes immutable and editable features, admissible education and occupation transitions, feasibility constraints, candidate generation rules, and the minimality objective used during neuro-symbolic counterfactual search.

\begin{lstlisting}[basicstyle=\ttfamily\scriptsize,
                   breaklines=true,
                   frame=single,
                   caption={Complete ASP program used by the PACE framework},
                   label={lst:asp}]
% Neuro-symbolic counterfactual search for the Adult Income dataset.
%
% This ASP file contains the symbolic knowledge base:
% - which features are editable or fixed
% - which changes are legal
% - which values are realistic
% - how to minimize the number of changed features
 
% ---------------------------------------------------------------------
% Editable and fixed features
% ---------------------------------------------------------------------
 
editable(hours).
editable(education).
editable(occupation).
 
fixed(age).
fixed(sex).
 
% ---------------------------------------------------------------------
% Legal value domains
% ---------------------------------------------------------------------
 
% Weekly hours can only move in small steps.
delta_hours(-10; -5; 0; 5; 10).
 
% Occupation changes are also local transitions.
can_change_occupation("Adm-clerical", "Exec-managerial").
can_change_occupation("Exec-managerial", "Adm-clerical").
can_change_occupation("Adm-clerical", "Sales").
can_change_occupation("Sales", "Adm-clerical").
can_change_occupation("Sales", "Tech-support").
can_change_occupation("Tech-support", "Sales").
can_change_occupation("Tech-support", "Prof-specialty").
can_change_occupation("Prof-specialty", "Tech-support").
can_change_occupation("Prof-specialty", "Exec-managerial").
can_change_occupation("Exec-managerial", "Prof-specialty").
can_change_occupation("Craft-repair", "Machine-op-inspct").
can_change_occupation("Machine-op-inspct", "Craft-repair").
can_change_occupation("Craft-repair", "Transport-moving").
can_change_occupation("Transport-moving", "Craft-repair").
can_change_occupation("Craft-repair", "Farming-fishing").
can_change_occupation("Farming-fishing", "Craft-repair").
can_change_occupation("Handlers-cleaners", "Other-service").
can_change_occupation("Other-service", "Handlers-cleaners").
can_change_occupation("Other-service", "Priv-house-serv").
can_change_occupation("Priv-house-serv", "Other-service").
can_change_occupation("Protective-serv", "Armed-Forces").
can_change_occupation("Armed-Forces", "Protective-serv").
 
% Education changes are local transitions only.
% This prevents unrealistic jumps such as Bachelors -> Doctorate.
can_change_education("Preschool", "1st-4th").
can_change_education("1st-4th", "Preschool").
can_change_education("1st-4th", "5th-6th").
can_change_education("5th-6th", "1st-4th").
can_change_education("5th-6th", "7th-8th").
can_change_education("7th-8th", "5th-6th").
can_change_education("7th-8th", "9th").
can_change_education("9th", "7th-8th").
can_change_education("9th", "10th").
can_change_education("10th", "9th").
can_change_education("10th", "11th").
can_change_education("11th", "10th").
can_change_education("11th", "12th").
can_change_education("12th", "11th").
can_change_education("12th", "HS-grad").
can_change_education("HS-grad", "12th").
can_change_education("HS-grad", "Some-college").
can_change_education("Some-college", "HS-grad").
can_change_education("Some-college", "Assoc-voc").
can_change_education("Assoc-voc", "Some-college").
can_change_education("Assoc-voc", "Assoc-acdm").
can_change_education("Assoc-acdm", "Assoc-voc").
can_change_education("Assoc-acdm", "Bachelors").
can_change_education("Bachelors", "Assoc-acdm").
can_change_education("Bachelors", "Masters").
can_change_education("Masters", "Bachelors").
can_change_education("Masters", "Prof-school").
can_change_education("Prof-school", "Masters").
can_change_education("Prof-school", "Doctorate").
can_change_education("Doctorate", "Prof-school").
 
% ---------------------------------------------------------------------
% Candidate generation
% ---------------------------------------------------------------------
 
1 { choose_hours(Delta) : delta_hours(Delta) } 1.
 
% Hours move in small steps and are clamped to the valid range [1, 80].
raw_hours(Raw) :- old_value(hours, Hours), choose_hours(Delta), Raw = Hours + Delta.
new_value(hours, 1)   :- raw_hours(Raw), Raw < 1.
new_value(hours, 80)  :- raw_hours(Raw), Raw > 80.
new_value(hours, Raw) :- raw_hours(Raw), Raw >= 1, Raw <= 80.
 
new_value(education, Education) :-
    old_value(education, Education),
    not changed_education.
 
0 { changed_education } 1.
 
1 { new_value(education, NewEducation) :
    can_change_education(Education, NewEducation) } 1 :-
    old_value(education, Education),
    editable(education),
    changed_education.
 
new_value(occupation, Occupation) :-
    old_value(occupation, Occupation),
    not changed_occupation.
 
0 { changed_occupation } 1.
 
1 { new_value(occupation, NewOccupation) :
    can_change_occupation(Occupation, NewOccupation) } 1 :-
    old_value(occupation, Occupation),
    editable(occupation),
    changed_occupation.
 
new_value(Feature, Value) :-
    old_value(Feature, Value),
    fixed(Feature).
 
% ---------------------------------------------------------------------
% Validity constraints
% ---------------------------------------------------------------------
 
% Fixed features cannot change.
:- fixed(Feature),
   old_value(Feature, Old),
   new_value(Feature, New),
   Old != New.
 
% Hours are kept in range by the clamping rules above.
% Earlier variant used for the Table 1 comparison discarded out-of-range
% hours instead of clamping; the two agree except at the hours boundaries:
% :- new_value(hours, Hours), Hours < 1.
% :- new_value(hours, Hours), Hours > 80.
 
% Changed occupations must follow the allowed transition graph.
:- old_value(occupation, Old),
   new_value(occupation, New),
   Old != New,
   not can_change_occupation(Old, New).
 
% ---------------------------------------------------------------------
% Change counting and minimality
% ---------------------------------------------------------------------
 
changed(hours) :-
    old_value(hours, Old),
    new_value(hours, New),
    Old != New.
 
changed(education) :-
    old_value(education, Old),
    new_value(education, New),
    Old != New.
 
changed(occupation) :-
    old_value(occupation, Old),
    new_value(occupation, New),
    Old != New.
 
% ASP ensures that generated candidates respect the budget.
:- change_budget(Budget),
   Budget != #count { Feature : changed(Feature) }.
 
#minimize { 1, Feature : changed(Feature) }.
 
#show new_value/2.
#show changed/1.
\end{lstlisting}
 
\section{Online Resources}
 
The source code is available via
\begin{itemize}
\item \href{https://github.com/PavelIakovets/PACE-A-Neuro-Symbolic-Framework-for-Plausible-and-Actionable-Counterfactual-Explanations}{GitHub}
\end{itemize}

\end{document}